\documentclass{article}
\usepackage{spconf,amsmath,graphicx}
\usepackage{enumitem}
\usepackage{adjustbox}
\usepackage{subcaption}
\usepackage{graphicx}
\usepackage{amsmath}
\usepackage{amssymb}
\usepackage{booktabs}
\usepackage{pifont}
\usepackage{comment}

    \title{Vista-Morph: Unsupervised Image Registration \\ of Visible-Thermal Facial Pairs}
\name{Catherine Ordun$^{\star,\dagger}$ \qquad Edward Raff $^{\star,\dagger}$ \qquad Sanjay Purushotham$^{\dagger}$}
\address{$^{\dagger}$University of Maryland, Baltimore County\\$^{\star}$Booz Allen Hamilton}

\begin{document}
%
\maketitle

\begin{abstract}
For a variety of biometric cross-spectral tasks, Visible-Thermal (VT) facial pairs are used. However, due to a lack of calibration in the lab, photographic capture between two different sensors leads to severely misaligned pairs that can lead to poor results for person re-identification and generative AI. To solve this problem, we introduce our approach for VT image registration called \textbf{Vista Morph}. Unlike existing VT facial registration that requires manual, hand-crafted features for pixel matching and/or a supervised thermal reference, Vista Morph is completely unsupervised without the need for a reference. By learning the affine matrix through a Vision Transformer (ViT)-based Spatial Transformer Network (STN) and Generative Adversarial Networks (GAN), Vista Morph successfully aligns facial and non-facial VT images. Our approach learns warps in Hard, No, and Low-light visual settings and is robust to geometric perturbations and erasure at test time. We conduct a downstream generative AI task to show that registering training data with Vista Morph improves subject identity of generated thermal faces when performing V2T image translation.

\end{abstract}

\begin{figure}[t!]
\centering
    \begin{subfigure}[b]{0.45\textwidth}
        \includegraphics[width=\textwidth]{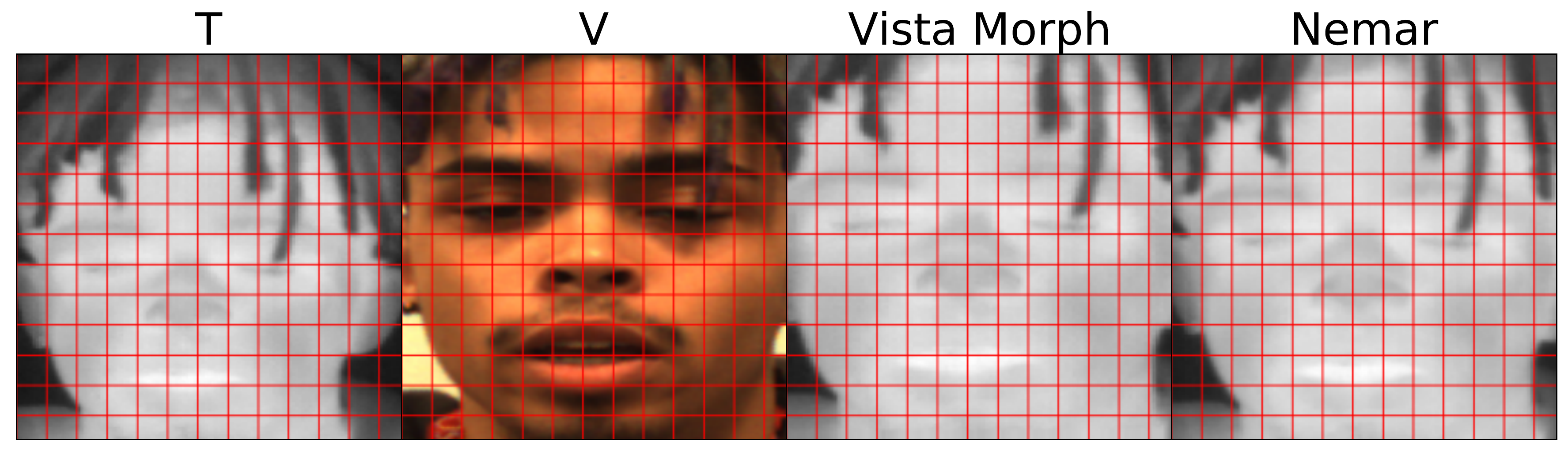}
        \caption{}
     \end{subfigure}  
    \begin{subfigure}[b]{0.45\textwidth}
    \centering
    \includegraphics[width=\textwidth]{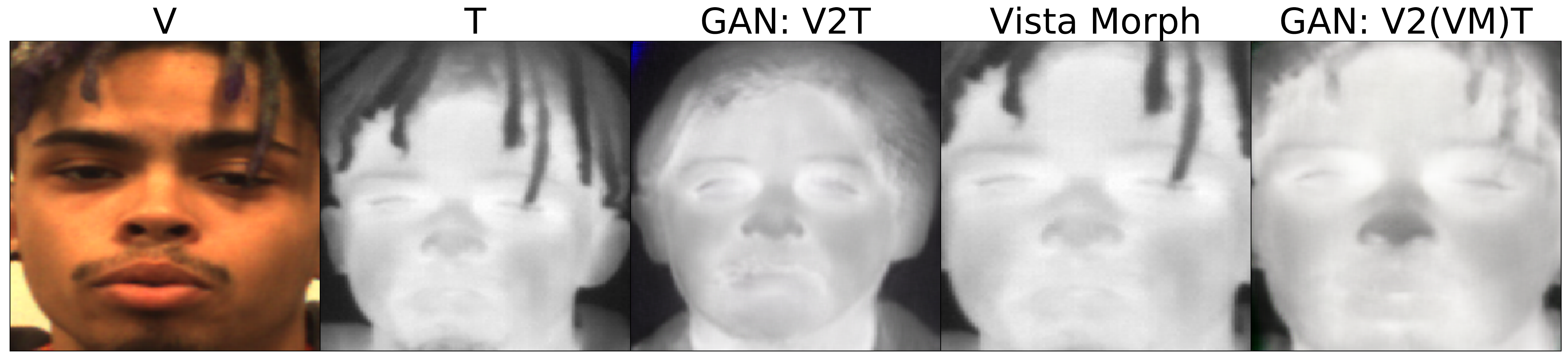}
    \caption{}
     \end{subfigure}  
    \begin{subfigure}[b]{0.45\textwidth}
    \centering
    \includegraphics[width=\textwidth]{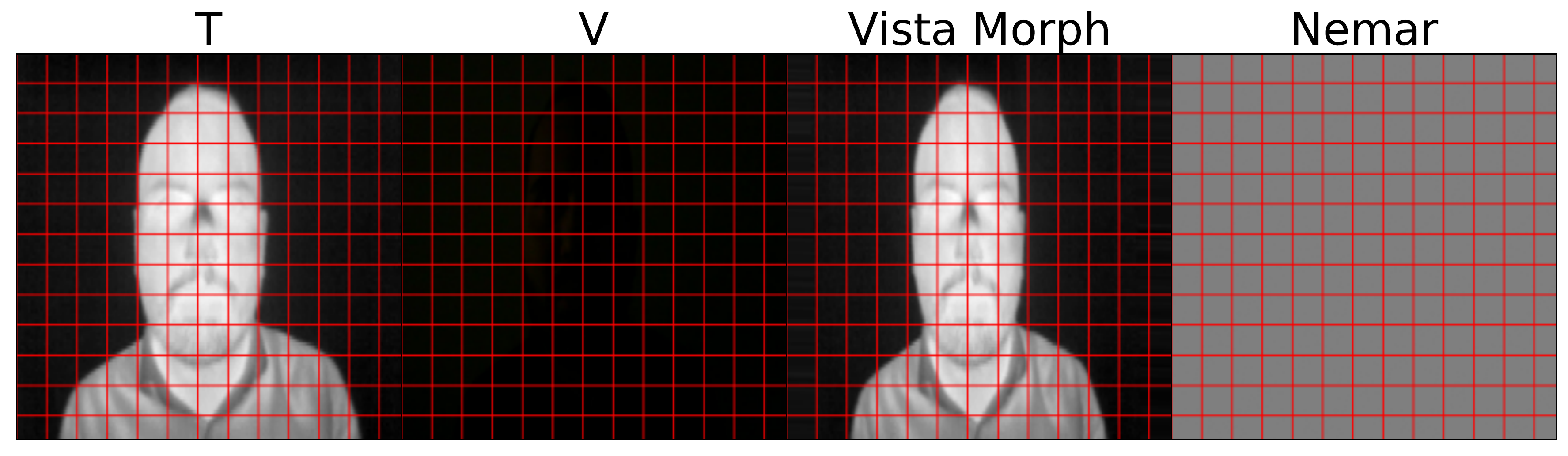}
    \caption{}
     \end{subfigure}
    \begin{subfigure}[b]{0.45\textwidth}
    \centering
    \includegraphics[width=\textwidth]{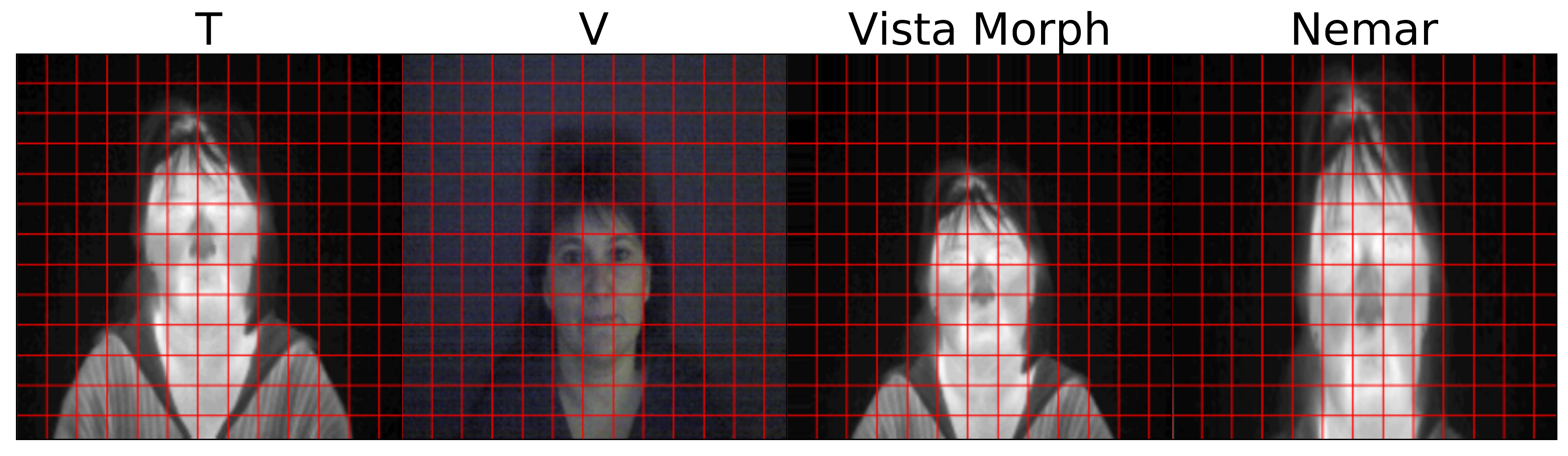}
    \caption{}
     \end{subfigure}  
    \caption{\textbf{Vista Morph Registered Samples.} \{a\} Vista Morph aligns \textbf{T}hermal faces relative to the scale of the \textbf{V}isible face more accurately than the existing state-of-the-art (Nemar). \{b\} Registration improves generated thermal identity when using a GAN. \{c,d\} Vista Morph registers thermal faces in No- and Low-Light visible settings better than Nemar.} 
    \label{main}
\end{figure}


\section{Introduction}
Multiple Visible-Thermal (VT) facial datasets are available for biometric tasks like emotion recognition, thermal face recognition, and person re-identification \cite{ordun2020use}. Unfortunately, misalignment is introduced at the time of data capture when two sensors (a thermal and visible camera) are positioned at different angles and distances. Given increasing interest in generative AI, this inherent misalignment between cross-spectral faces can weaken image quality in generative tasks \cite{kong2021breaking} such as Visible-to-Thermal (V2T) image translation, due to shift invariance \cite{zhang2019making}.  Manual scaling, cropping, and alignment by hand is infeasible when dealing with thousands of images. Further, existing VT alignment methods rely on supervised feature matching \cite{kong2007multiscale,krishnan2022intensity,ma2015non,sun2017thermal}. As a result, to rapidly register VT faces on multiple VT facial datasets of varying scale and distortion, we offer \textbf{Vis}ible \textbf{T}hermal Facial Morph (\emph{Vista Morph}). Our model is the first unsupervised approach to register VT faces, to our knowledge, and does not rely on feature matching or a target reference. Vista Morph combines two Generative Adversarial Networks (GAN) \cite{goodfellow2014generative} and a Spatial Transformer Network (STN) \cite{jaderberg2015spatial} that uses Vision Transformer (ViT) \cite{dosovitskiy2020image} for the first time, as the localization network. This contrasts from similar cross-spectral/multi-modal works \cite{benvenuto2022unsupervised,deng2009imagenet,upendra2021joint,zhang2021single} that rely on a traditional CNN or U-NET \cite{ronneberger2015u} localization network for the STN. We select ViT because it applies self-attention across embedded image patches, making the spatial information fixed and preserved across layers of the network, whereas CNNs are less spatially discriminative \cite{mishra2021vt,phan2022patch,raghu2021vision}. Unlike traditional image registration methods, no similarity metric such as mean squared difference, normalized cross-correlation, or mutual information is optimized during training \cite{de2019deep}. Only common GAN-based losses are learned. Further, Vista Morph integrates a Fourier Loss to learn how to align thermal images relative to No- and Low-light visible pairs by relying on the signal domain - so far unexplored in VT image registration but critical since Long-Wave Infrared (LWIR) sensors capture visible faces without the need for a light source.

We evaluate three VT facial datasets to align thermal faces relative to the visible face's geometry (T$\sim$V) and vice-versa (V$\sim$T). To examine generative image quality, we register each dataset with Vista Morph and train a conditional GAN \cite{ordun2023visible} for the downstream task of Visible-to-Thermal (V2T) image translation. V2T image translation is increasingly researched for its value in person re-identification, thermal face recognition, and thermal physiology \cite{chen2019matching,hermosilla2021thermal,kniaz2018thermalgan,merla2004emotion,nair2022t2v,pavlidis2007interacting,pavlidis2000imaging,zhang2018tv}. We also train a Diffusion Model for the T2V generative task \cite{ho2020denoising,nichol2021improved}. We then use diagrams of the underlying facial vasculature, a thermal biometric asserted by \cite{buddharaju2007physiology}, to analyze similarity between real and generated thermal identities. Our paper ends with a series of ablation studies on architectural settings and robustness. Our contributions are the following: 

\begin{itemize}[noitemsep]
    \item The first unsupervised VT facial image registration called Vista Morph that uses ViT, for the first time, as a localization network in the STN framework. 
    \item Registering pairs in challenging No- and Low-Light settings, a common scenario when using thermal sensors, by integrating a Fourier Loss in the Vista Morph model.
    \item Analyzing the identity of \emph{generated} thermal faces from GANs by extracting vessel maps that visualize underlying thermal vasculature. 
    \item Generalizability beyond faces with Vista Morph application to automated driving datasets and proven robustness against geometric transformations and erasure.
\end{itemize}

\begin{figure*}[ht!]
\centering
    \begin{subfigure}[b]{0.27\textwidth}
        \includegraphics[width=\textwidth]{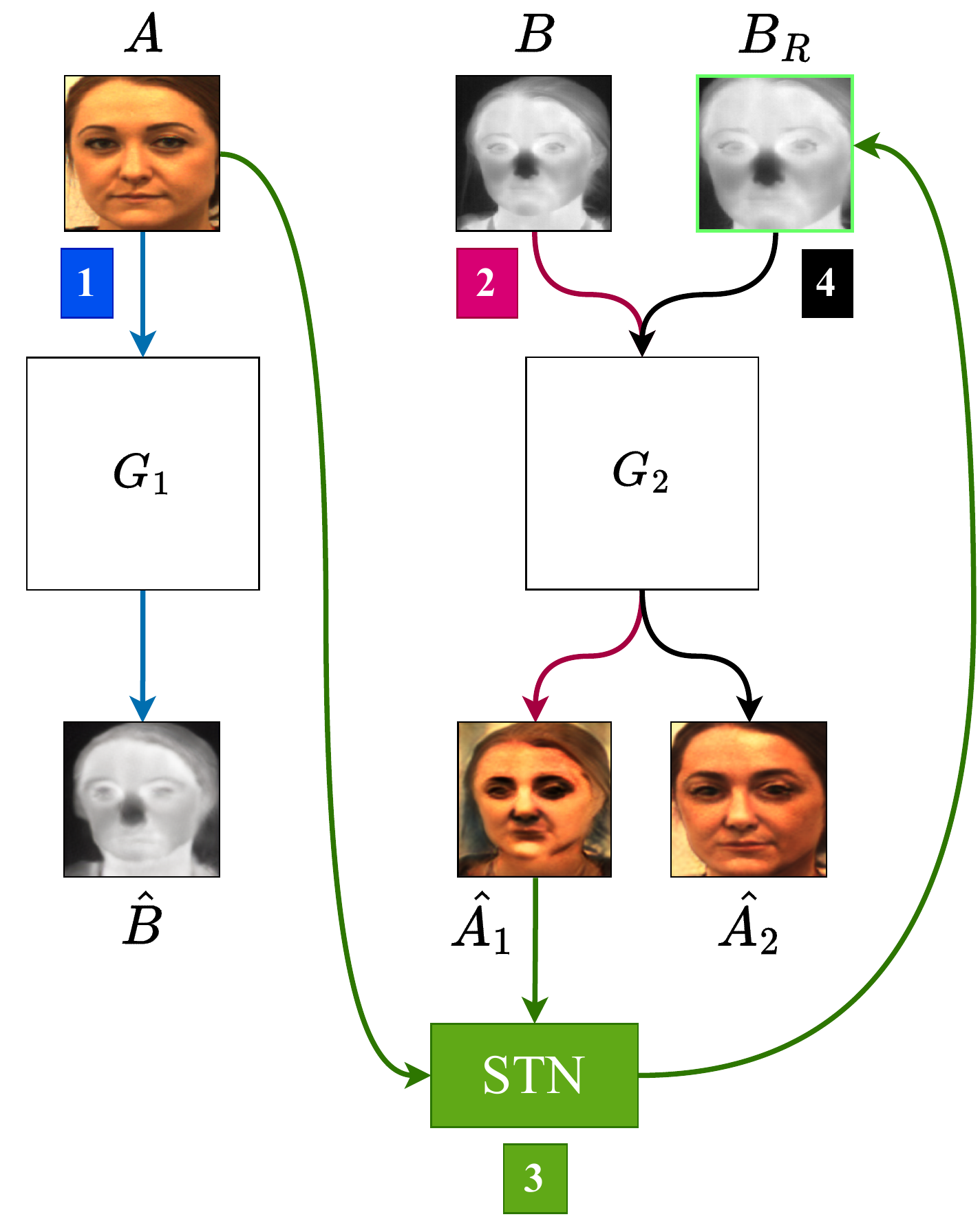}
        \caption{Flows}
        \label{flows}
     \end{subfigure}    
     \begin{subfigure}[b]{0.4\textwidth}
         \includegraphics[width=\textwidth]{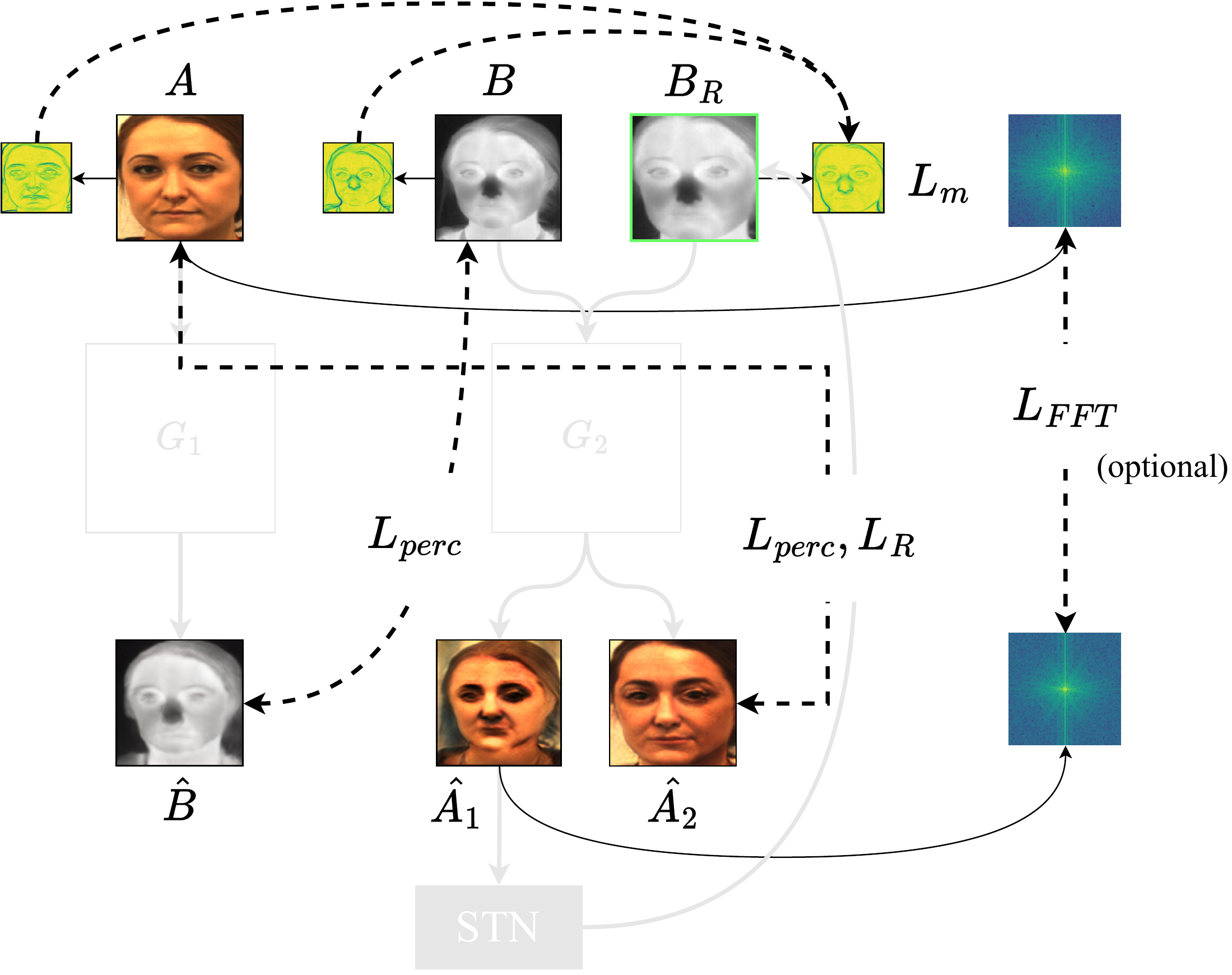}
         \caption{Losses}
         \label{losses}
     \end{subfigure}
    \begin{subfigure}[b]{0.3\textwidth}
        \includegraphics[width=\textwidth]{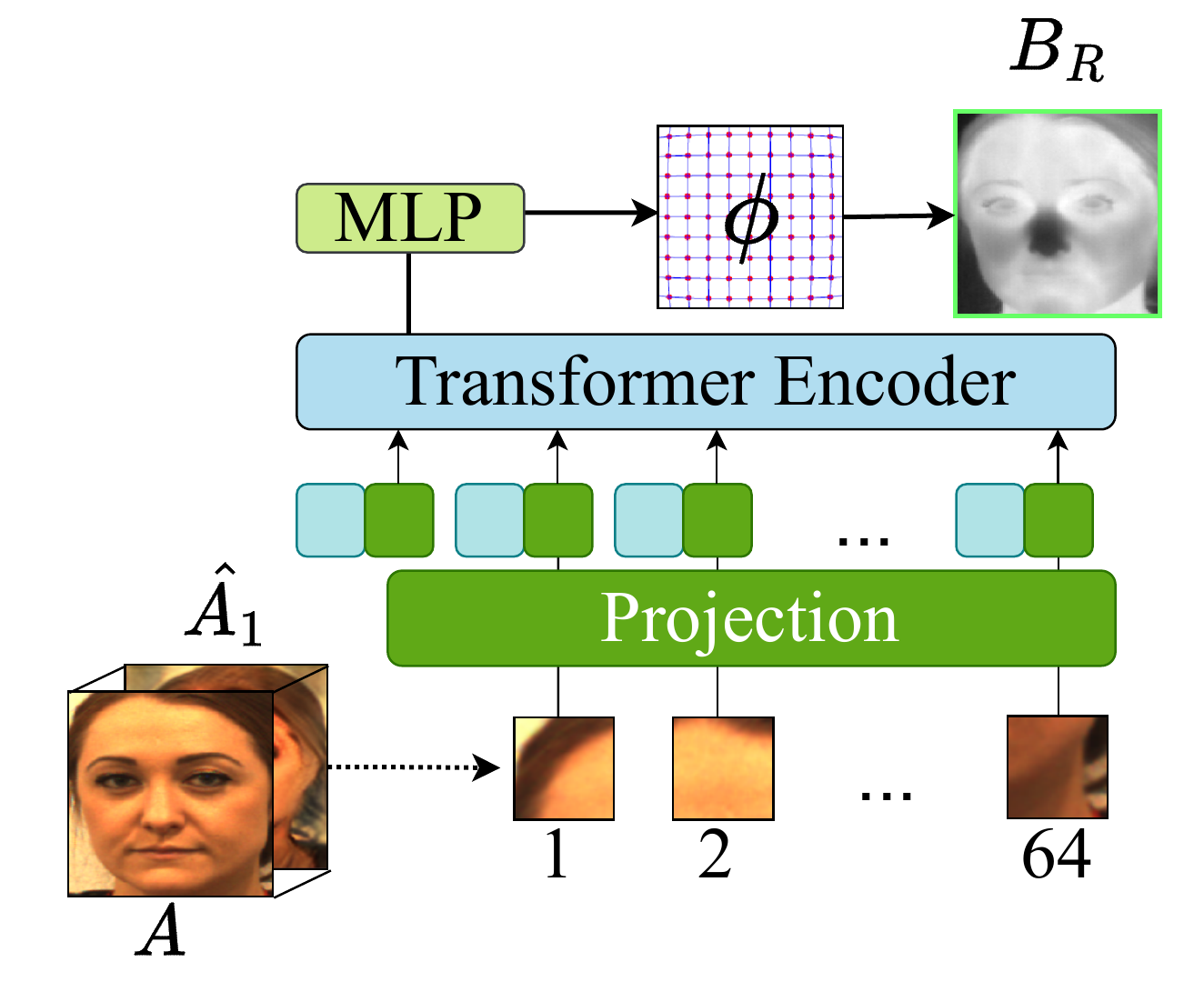}
        \caption{STN with ViT}
        \label{stn}
     \end{subfigure}    
    \caption{\textbf{Training Overview.} Four flows trained in an end-to-end manner pass through a STN designed using a ViT. This process outputs the affine matrix ($\theta)$ used to warp the real thermal image ($B$) to its registered form ($B_R$) based on the intermediate fake visible image ($\hat{A_2}$). By translating $B \rightarrow \hat{A_1}$ using $G_2$, the STN can learn the scale between the visible and thermal spectra, using $\hat{A_1}$ as a proxy for $B$, all the while ensuring that $B_R$ generates a visible face $\hat{A_2}$ similar to the original $A$.}
    \label{vistamorph}
\end{figure*}

\section{Related Works}
Existing VT facial registration relies on feature-based matching such as edge maps, corner detection, intensity histograms, and SIFT features, or a supervised target, where these methods only evaluate on a single VT face dataset \cite{kong2007multiscale,krishnan2022intensity,ma2015non,sun2017thermal}. Multimodal medical image registration methods such as DLIR \cite{de2019deep} and Voxelmorph \cite{balakrishnan2019voxelmorph} are applied for CT and MRI imagery. However, while the images vary in density, they are still captured in the same optical spectra. This differs from our challenge where two images are obtained in different electromagnetic spectra altogether; the visible band (350 - 740 nm) and the LWIR band (8 - 15 $\mu$m). The task of unsupervised cross-spectral image translation is new. The most similar work to ours is the sentinel Nemar algorithm by Arar et al., \cite{arar2020unsupervised} that first demonstrated unsupervised VT image registration on non-facial images. Since then, several similar Nemar-like approaches to CT/MRI images, remote sensing, and VT street scenes have been developed using varying translation and registration flows, new loss functions and/or fusion \cite{chen2022unsupervised,kong2021breaking,wang2022unsupervised,xu2022nbr}. No existing works tackle the challenge of non-rigid cross-spectral facial images since they contain abrupt changes and sudden deformations.

\section{Vista Morph}
We describe our approach, Vista Morph, by describing the training flows, loss functions to include the Fourier Loss for handling No- and Low-light settings, as well as using ViT as a novel localization network with a custom Multilayer Perceptron (MLP) for the regressor network of the STN \cite{jaderberg2015spatial} framework. 

\subsection{Generative Flows}
Shown in Figure \ref{flows}, registration is trained using four flows, in an end-to-end fashion. First, the ground-truth visible image, $A$, is passed to the first Generator, $G_1$, that outputs the fake thermal image, $\hat{B}$. Second, the original thermal image, $B$, is passed to the second Generator, $G_2$, that outputs the first fake visible image, $\hat{A_1}$. Third, both $A$ and $\hat{A_1}$ are used as inputs to the STN, in order to output the registered thermal image, $B_R$. In the fourth flow, $B_R$ is passed back to $G_2$ in order to output $\hat{A_2}$. These flows force the STN to use a visible mapping of $\hat{B}$ translated into its thermal counterpart, $\hat{A_1}$, that preserves the geometry of the original thermal image as indicated in Figure \ref{stn}. By translating the visible to the thermal spectrum, the STN can now learn the scale between both modalities. Two GANs are used, where the generators, $G1$ and $G2$, are identical U-NETs with 5 encoder and 4 decoder modules with added BlurPool layers \cite{zhang2019making}. The discriminators, $D1$ and $D2$ are identical and comprise a traditional PatchGAN \cite{isola2017image} architecture with a 16x16 patch, also incorporating BlurPool layers.  

\setlength{\belowdisplayskip}{-2pt} 
\setlength{\abovedisplayskip}{-3pt}

\textbf{GAN Losses.} The perceptual quality of the VT images is controlled using an LPIPS \cite{zhang2018unreasonable} perceptual loss, $L_{perc}$. Per Eq. \ref{eq1}, $\phi$ is the VGG-16 network and $\tau$ transforms network embeddings.

\begin{equation} \label{eq1}
\begin{split}
L_{\mathit{perc}} = \sum_{n}\tau^n(\phi^n(\hat{B}) - \phi^n(B)) + \sum_{n}\tau^n(\phi^n(\hat{A}) - \phi^n(\hat{A_2}))
\end{split}
\end{equation}

Recall that $\hat{A_2}$ is the visible image outputted by $B_R$ after being warped by the STN. To enforce the alignment of $B_R$ against the desired visible geometry of $A$, we set an L1 reconstruction loss shown in Eq. \ref{eq2}.

\begin{equation} \label{eq2}
L_R = \| A - \hat{A_2}\|_1
\end{equation}

To control for structural similarity between $B_R$ and $A$, we calculate morphological gradients for $B_R$, $B$, and $A$, and apply a triplet loss in Eq. \ref{eq3}. 

\begin{equation} \label{eq3}
\begin{split}
L_{\mathit{m}} = \frac{1}{K}\sum_{k=1}^{K} \max \{d(B_R, A) - d(B_R, B) + {\rm 1}, 0\}
\end{split}
\end{equation}

Finally, for datasets with Low- or No-Light visible imagery, we add a \textbf{Fourier Loss} to learn the signal domain, as opposed to only spatial domain. The $L_{FFT}$ shown in Eq. \ref{fft} is the L1 loss of the amplitude in Eq. \ref{eq_AMP_} and phase in Eq. \ref{eq_PHA_} of $A$ and $\hat{A1}$. 

\setlength{\belowdisplayskip}{-3pt} 
\setlength{\belowdisplayshortskip}{0pt}
\setlength{\abovedisplayshortskip}{-1pt}

\begin{equation} \label{eq_AMP_}
L_{amp}= \| (\lvert\mathcal{F}\{A\}_{u,v}\rvert) -(\lvert\mathcal{F}\{\hat{A_1}\}_{u,v}\rvert) \|_1
\end{equation}

\begin{equation} \label{eq_PHA_}
L_{pha} = \| (\angle{\mathcal{F}}\{A\}_{u,v}) -(\angle{\mathcal{F}}\{\hat{A_1}\}_{u,v})\|_1
\end{equation}

\begin{equation} \label{fft}
L_{FFT} = L_{amp}(A, \hat{A_1}) + L_{pha}(A, \hat{A_1})
\end{equation}

We train the GAN ($L_{GAN}$) using a relativistic adversarial loss \cite{jolicoeur2018relativistic} leading to the total Generator Loss, $L_G$ shown in Equation \ref{eq4}. The total Discriminator loss $L_D$, is an average of the real and fake discriminator losses which are both relativistic. The total training objective is shown in Equation \ref{eq5}. 

\setlength{\belowdisplayskip}{-5pt} 
\setlength{\abovedisplayskip}{-5pt}

\begin{equation} \label{eq4}
\begin{split}
L_{G} = L_{GAN} + L_{perc} + L_{R} + L_{m}
\end{split}
\end{equation}

\begin{equation} \label{eq5}
G^* = \arg \min_G \max_D L_{G} + L_{D} 
\end{equation}

\subsection{Registration Network}
The registration network is a STN \cite{jaderberg2015spatial}. STN is not a model in itself, but rather, a framework where any differentiable function (e.g. neural network) can be used as the localization network. As a result, we use a 12-Layer ViT \cite{dosovitskiy2020image} as the localization network to extract features between the visible (aligned) and thermal (non-aligned) image, and add a 4-Layer MLP as the regressor network. Shown in Figure \ref{stn}, the concatenated input of $(A, \hat{A_1})$ are passed to the ViT using a patch size of 64. The MLP consists of the following architecture: Linear (17*768,1024)-ReLU-Linear (1024,512)-Relu-Linear (512,256)-Sigmoid-Linear (256,6), outputting an affine matrix ($\phi$) of size 6. Using a sigmoid activation function is important for the regressor since it outputs values between [-1,1] in the range of $\phi$ \cite{jaderberg2015spatial}. The $\phi$ is calculated for each $(A,A_1)$ pair which represents the 2D flow field (sampling grid), given a batch of affine matrices. Using an affine transformation, the STN computes the registered output, $B_R$ sampling the pixel locations from the field. The STN is trained jointly with $G1$ and $G2$. 

\section{Experiments}
\subsection{Datasets} In this section, we evaluate our approach on three VT paired facial datasets: Carl \cite{espinosa2013new}, Army Research Lab (ARL) Devcom Dataset \cite{poster2021large}, and Eurecom \cite{mallat2020facial}. To test our approach on  a non-facial domain, we use the FLIR Advanced Driver Assistance Systems (ADAS) dataset \cite{flir_adas_readme}. For each dataset, we conduct a minimal amount of preprocessing. For example, with the Devcom dataset, we use the forward-facing ``baseline" and ``expression" protocols and ignore the thermal bounding box metadata supplied for alignment. Instead, we use a FaceNet MTCNN  \cite{lugaresi2019mediapipe,zhang2016joint} to detect and crop visible faces, and apply a series of binary thresholding operations to crop the thermal face image away from its background. The results lead to a misaligned set of VT facial pairs with varying degrees of warp. We select a random 5\% sample from our misaligned Devcom dataset (56,205 training pairs) due to compute limitations. We use the entire Eurecom, Carl, and FLIR ADAS datasets, with details in Table \ref{datasets}. 

\begin{figure}[ht]
\centering
\includegraphics[width=0.48\textwidth]{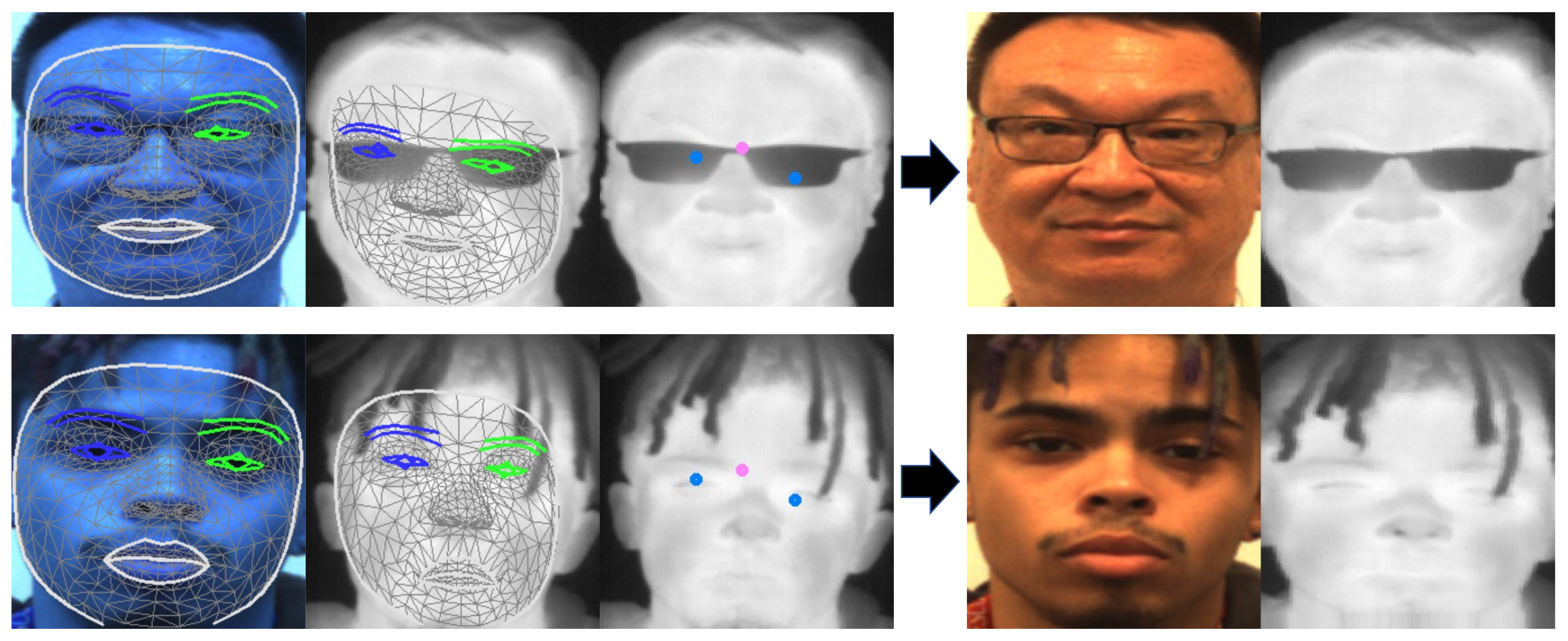}
\caption{\textbf{Manual Alignment.} Existing facial keypoint models work for visible spectra, but fail to consistently detect thermal facial landmarks. If available, manually estimating the affine matrix leads to misaligned pairs.}
\label{manual}
\end{figure}

\begin{figure*}[t]
\centering
     \begin{subfigure}[b]{0.45\textwidth}
         \includegraphics[width=\textwidth]{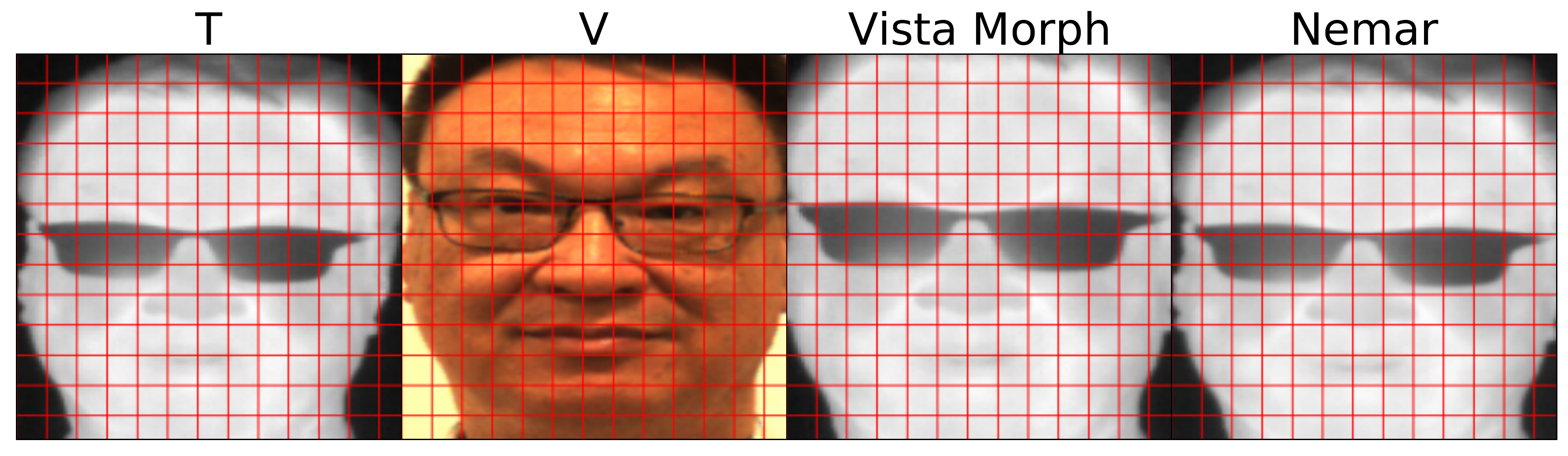}
         \caption{Devcom T$\sim$V}
         \label{a}
     \end{subfigure}   
     \begin{subfigure}[b]{0.45\textwidth}
         \includegraphics[width=\textwidth]{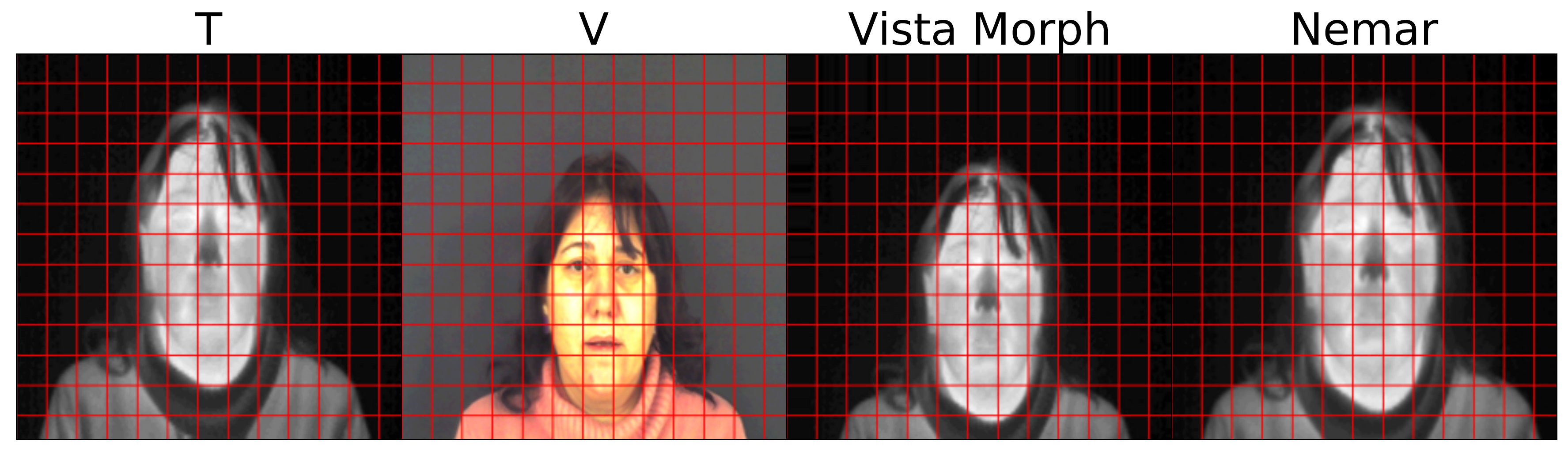}
         \caption{Carl T$\sim$V}
         \label{b}
     \end{subfigure}
    \begin{subfigure}[b]{0.45\textwidth}
        \includegraphics[width=\textwidth]{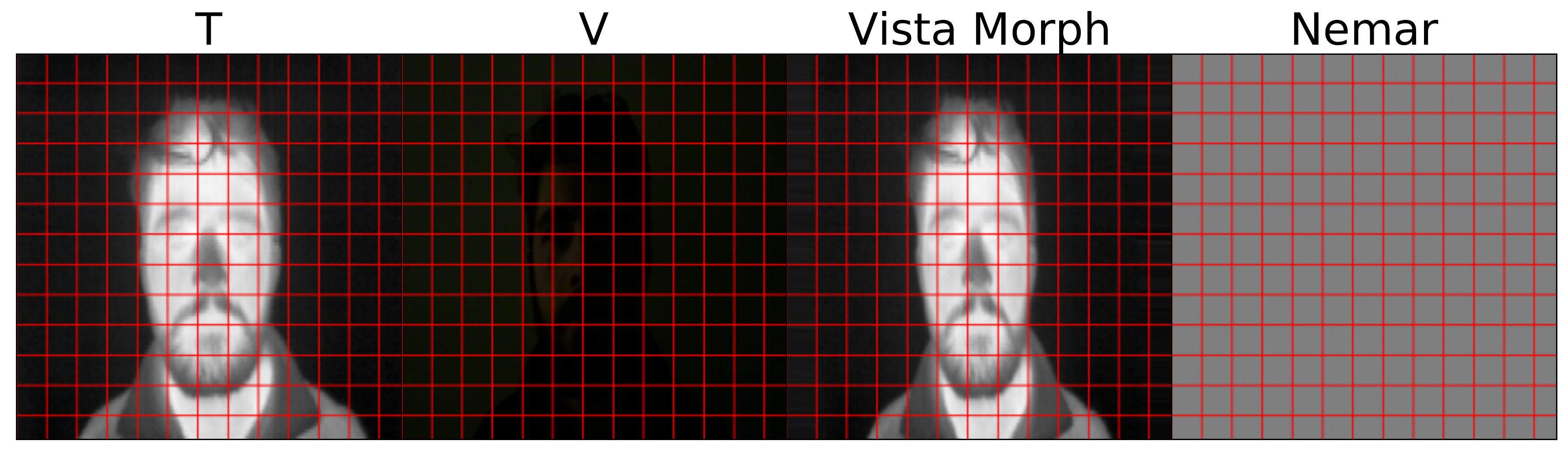}
        \caption{Eurecom T$\sim$V}
        \label{c}
     \end{subfigure}   
     \begin{subfigure}[b]{0.45\textwidth}
        \includegraphics[width=\textwidth]{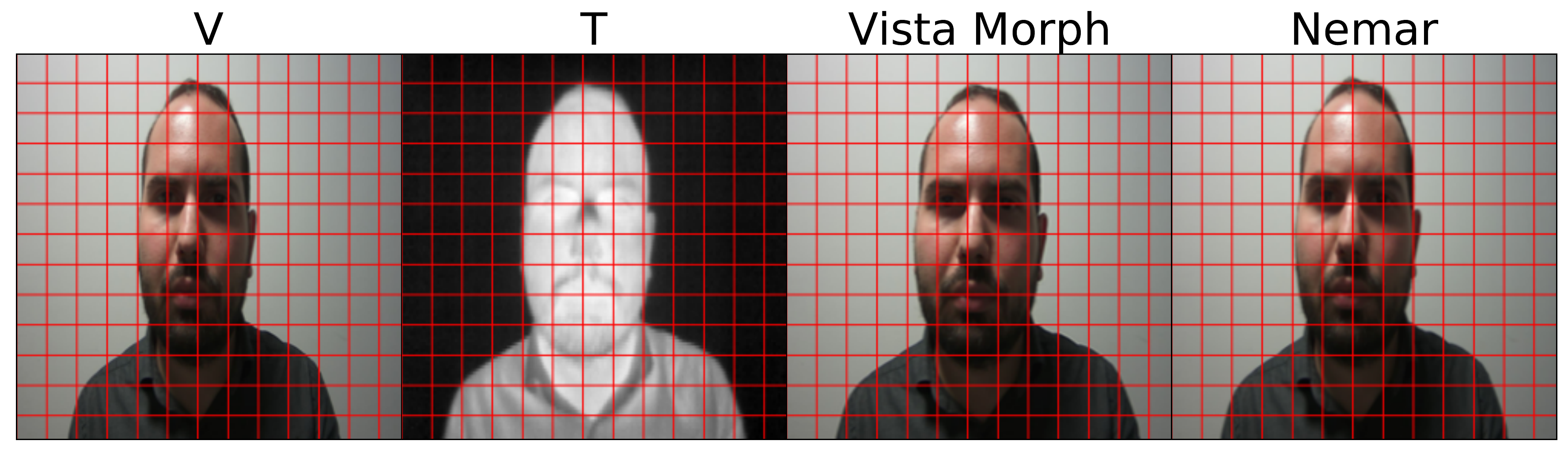}
        \caption{V$\sim$T}
        \label{d}
     \end{subfigure}
    \caption{\textbf{Registration Samples}.  Row 1: Devcom T$\sim$V, Row 2: Carl T$\sim$V, Row 3: Eurecom T$\sim$V, Row 4: Eurecom V$\sim$T. }
    \label{grid}
\end{figure*}

\begin{figure*}[t]
\centering 
    \begin{subfigure}[b]{0.9\textwidth}
        \includegraphics[width=\textwidth]{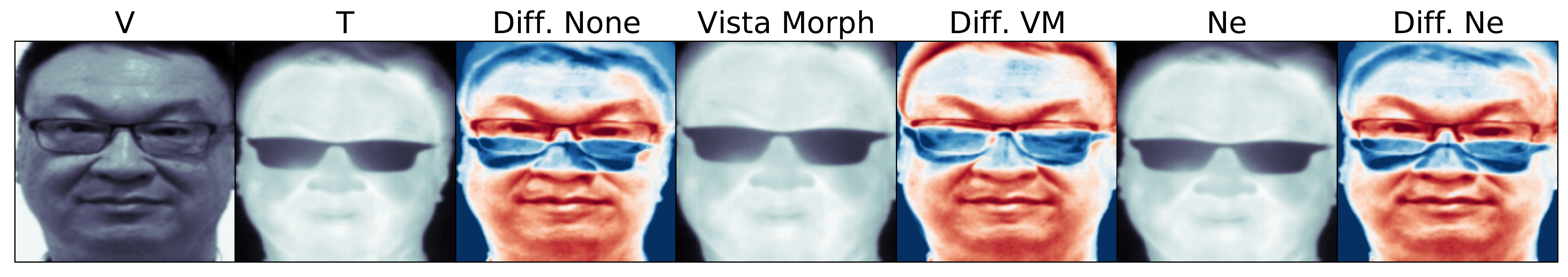}
     \end{subfigure}   
    \begin{subfigure}[b]{\textwidth}
    \centering
        \includegraphics[width=0.9\textwidth]{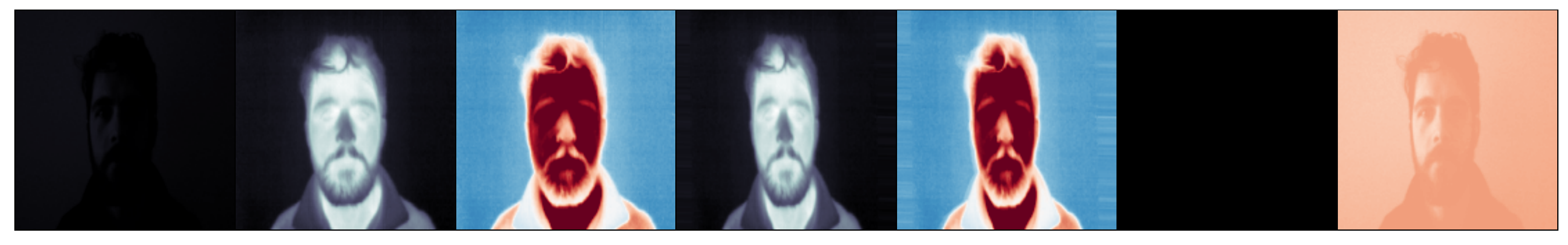}
     \end{subfigure}    
    \caption{\textbf{Difference Maps Before and After Registration}. The ``Diff.VM" heatmap displays the most consistent overlap between red (visible) and blue (thermal) color maps. Notice the precise placement of sunglasses, compared to ``Diff.Ne". Vista Morph registers the thermal image even when the visible image is No-Light, making apparent the ``Diff.VM" heatmap. The ``Diff.Ne" heatmap only shows the visible image in red because no thermal image was found/registered. V: Visible, T: Thermal, Diff. None (T-V), Diff. VM (Vista Morph - V), Diff. Ne (Ne - V)}
    \label{difference}
\end{figure*}

\begin{table}[ht]
\begin{adjustbox}{width=0.48\textwidth}
\begin{tabular}{@{}lllll@{}}
\toprule
\textbf{Dataset} & \textbf{Direction} & \textbf{Train Subj} & \textbf{Test Subj} & \textbf{Train Pairs} \\ \midrule
Devcom 5\%        & T $\sim$V & 376   & 74  & 2810    \\
Carl          & T $\sim$V & 32    & 9   & 1920    \\
Eurecom (Vis) & V $\sim$T & 45    & 5   & 945     \\
ADAS          & T $\sim$V & N/A   & N/A & 7,521   \\ \toprule
\textbf{Dataset} & \textbf{Direction} & \textbf{Test Pairs} & \textbf{Positions} & \textbf{Lighting}    \\ \midrule
Devcom 5\%       & T $\sim$V & 247   & F   & H       \\
Carl          & T $\sim$V & 540   & F   & H,L     \\
Eurecom (Vis)    & V $\sim$T          & 105                 & F, U, D, S         & H, L, N              \\
ADAS          & T $\sim$V & 1,159 & N/A & H, L, N \\ \bottomrule
\end{tabular}
\end{adjustbox}
\caption{\textbf{Datasets}. Positions: \textbf{F}rontal, \textbf{U}p, \textbf{D}own, \textbf{S}ideways. Lighting: \textbf{H}ard, \textbf{L}ow, \textbf{N}one.}
\label{datasets}
\end{table}

\begin{figure*}[htbp!]
\centering
     \begin{subfigure}[b]{0.9\textwidth}
        \centering
         \includegraphics[width=\textwidth]{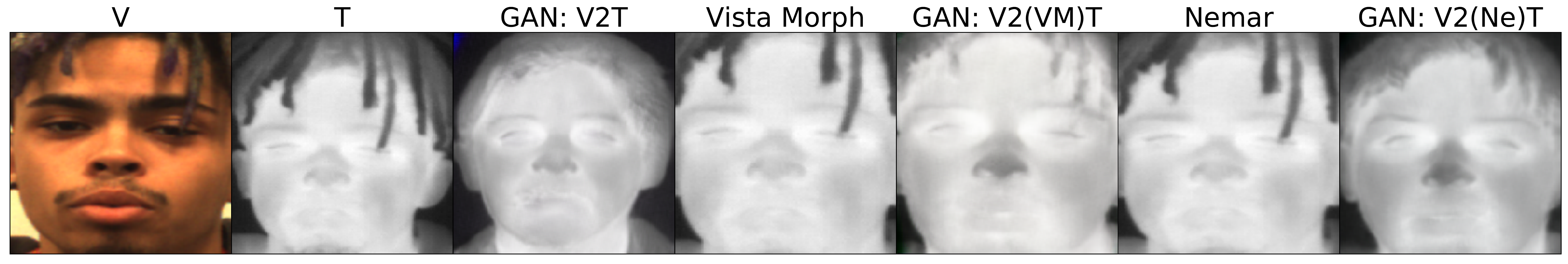}
     \end{subfigure}  
     \begin{subfigure}[b]{0.9\textwidth}
        \centering
        \includegraphics[width=\textwidth]{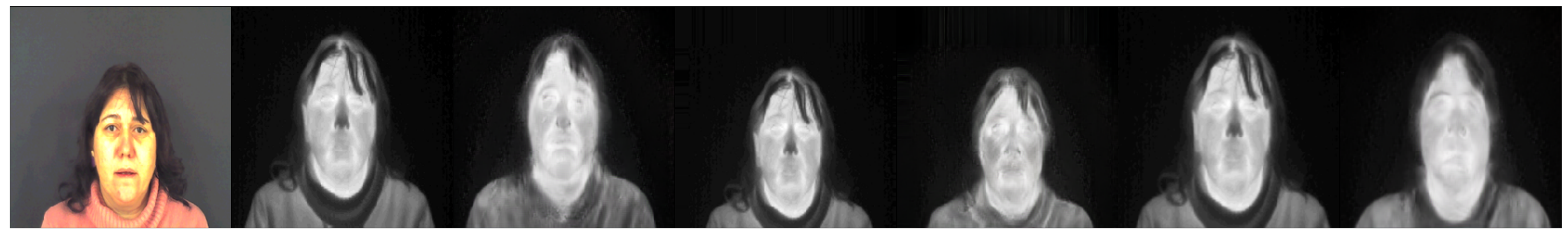}
     \end{subfigure}

    \begin{subfigure}[b]{0.9\textwidth}
        \centering
        \includegraphics[width=\textwidth]{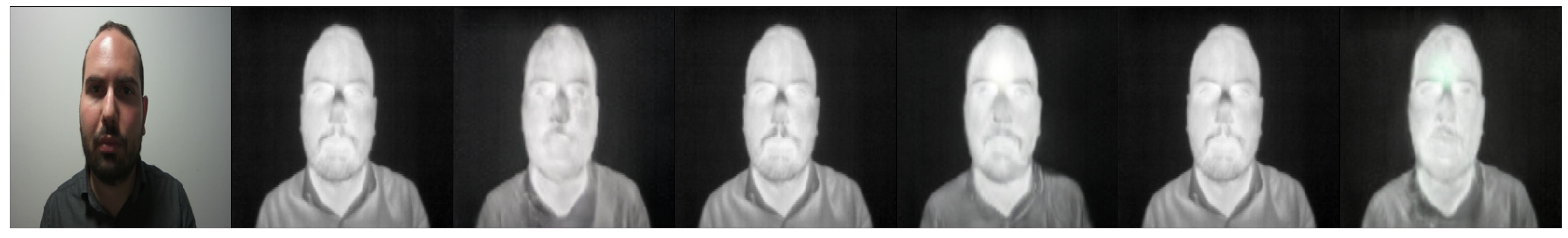}
     \end{subfigure}    
    \caption{\textbf{Generated Thermal Faces for V2T Image Translation.} The first two cols. are the original V and T faces. ``GAN: V2T" is the generated thermal face from VTF-GAN using unregistered pairs. ``Vista Morph" (VM) is the registered thermal face. ``GAN: V2(VM)T" is the generated thermal face when VTF-GAN is trained on VM-registered pairs. ``Nemar" is the registered thermal face using the Nemar baseline. ``GAN: V2(Ne)T" is the generated thermal face when VTF-GAN is trained on Nemar-registered pairs. Notice that for all datasets, the quality and identity of the ``GAN: V2(VM)T" has less distortion, artifacts, and more qualitative similarity to the registered thermal face.}
    \label{gan}
\end{figure*}

\begin{figure}[ht!]
\centering
\includegraphics[width=0.48\textwidth]{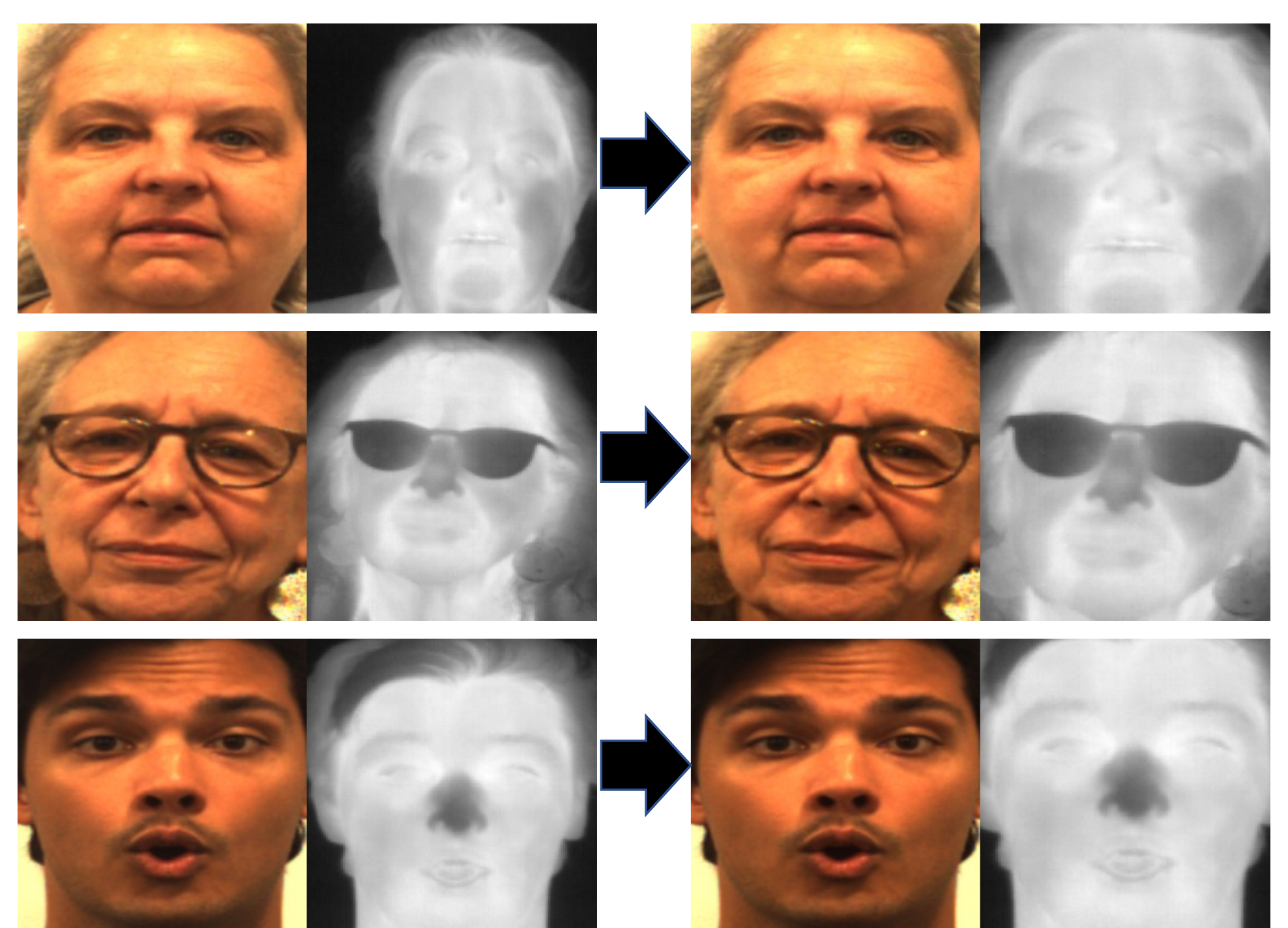}
\caption{\textbf{Vista Morph Results under Varying Warps and Scale.} Even when thermal faces are grossly misaligned (i.e. off-center, tilted with glasses), Vista Morph can learn the affine matrices to register $T\sim V$.}
\label{skewness}
\end{figure}

\begin{table}[]
\centering
\begin{adjustbox}{width=0.45\textwidth}
\begin{tabular}{@{}lllllr@{}}
\toprule
                 &                    & \multicolumn{4}{c}{\textbf{SSIM Edges} $\uparrow$}                                                              \\ \midrule
\textbf{Dataset} & \textbf{Alig.} & \textbf{No Reg.} & \textbf{VM} & \textbf{Ne} & \multicolumn{1}{l}{\textbf{Diff.}} \\ \midrule
Devcom           & T $\sim$V          & 0.878            & \textbf{0.898}      & 0.894          & \textbf{0.5\%}                                      \\
Carl             & T $\sim$V          & 0.804            & \textbf{0.861}      & 0.820          & \textbf{5.1\%}                             \\
Eurecom (Thr)    & T $\sim$V          & 0.833            & \textbf{0.837}      & 0.830          & \textbf{0.8\%}                                      \\ \midrule
                 &                    & \multicolumn{4}{c}{\textbf{NCC Edges} $\uparrow$}                                                               \\ \midrule
\textbf{Dataset} & \textbf{Alig.} & \textbf{No Reg.} & \textbf{VM} & \textbf{Ne} & \multicolumn{1}{l}{\textbf{Diff.}} \\ \midrule
Devcom           & T $\sim$V          & 0.013            & \textbf{0.147}      & 0.079          & \textbf{86.6\%}                                     \\
Carl             & T $\sim$V          & 0.191            & \textbf{0.411}      & 0.201          & \textbf{105.0\%}                           \\
Eurecom (Thr)    & T $\sim$V          & 0.285            & \textbf{0.329}      & 0.320          & \textbf{2.9\%}                             \\ \midrule
                 &                    & \multicolumn{4}{c}{\textbf{Mutual Information (MI) $\uparrow$}}                                                 \\ \midrule
\textbf{Dataset} & \textbf{Alig.} & \textbf{No Reg.} & \textbf{VM} & \textbf{Ne} & \multicolumn{1}{l}{\textbf{Diff.}} \\ \midrule
Devcom           & T $\sim$V          & 0.246            & 0.269               & \textbf{0.270} & -0.2\%                                     \\
Carl             & T $\sim$V          & 0.473            & \textbf{0.649}      & 0.512          & \textbf{26.8\%}                            \\
Eurecom (Thr)    & T $\sim$V          & 0.493            & \textbf{0.526}      & 0.474          & \textbf{11.0\%}                            \\ \bottomrule
\end{tabular}
\end{adjustbox}
\caption{\textbf{Registration Results.} Best scores are in \textbf{bold}. VM: Vista Morph, Ne: Nemar, Diff: Relative percentage change of VM over Ne. Bold Diff. scores indicate VM improvement over Ne.}
\label{Registration_Results}
\end{table}

\begin{figure*}[t]
\centering
\includegraphics[width=0.99\textwidth]{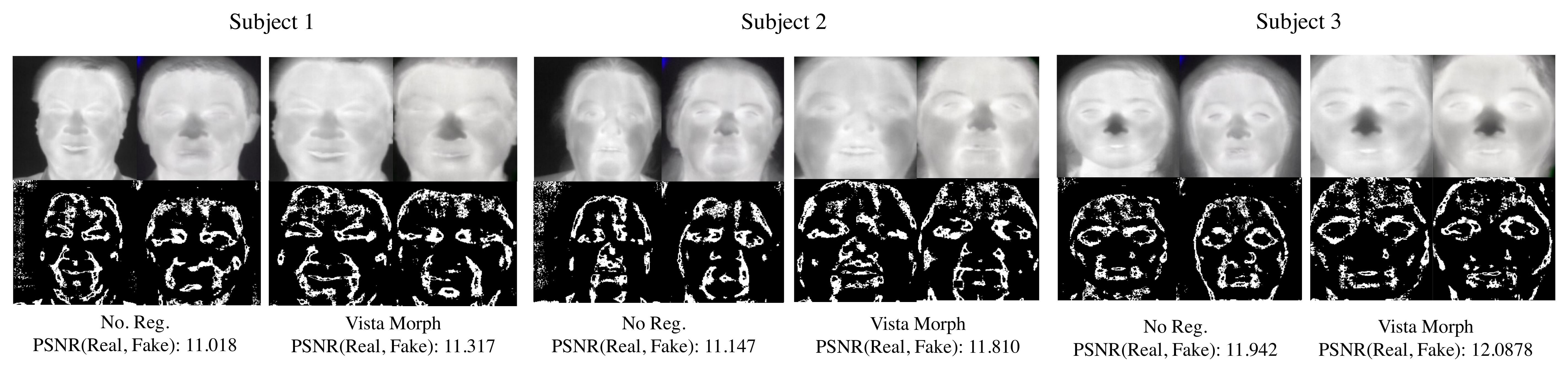}
\caption{\textbf{Thermal Vessels Before and After Vista Morph Registration.} Identity is more similar when VTF-GAN is trained on Vista Morph registered data.}
\label{vessels}
\end{figure*}

\begin{figure}[t]
\centering
     \begin{subfigure}[b]{0.3\textwidth}
        \centering
         \includegraphics[width=\textwidth]{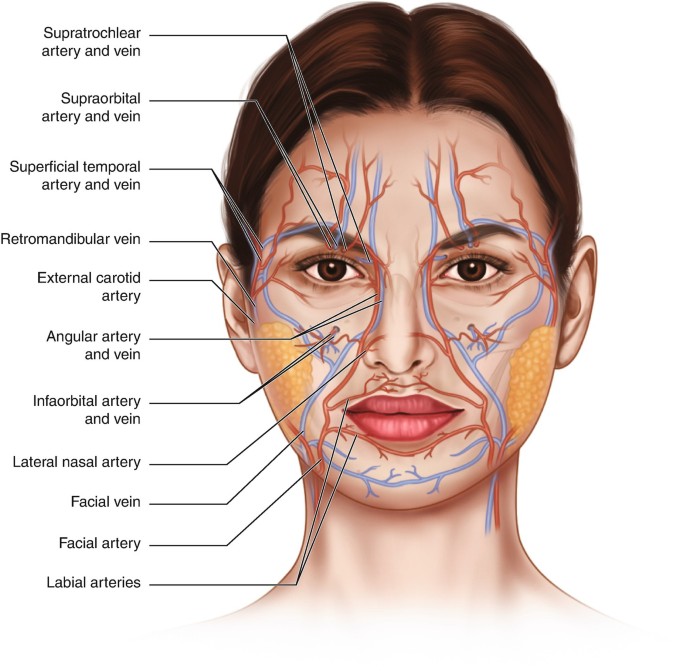}
         \caption{Facial anatomy of the face (Google Images).}
         \label{artery}
     \end{subfigure}  
     \begin{subfigure}[b]{0.45\textwidth}
        \centering
        \includegraphics[width=\textwidth]{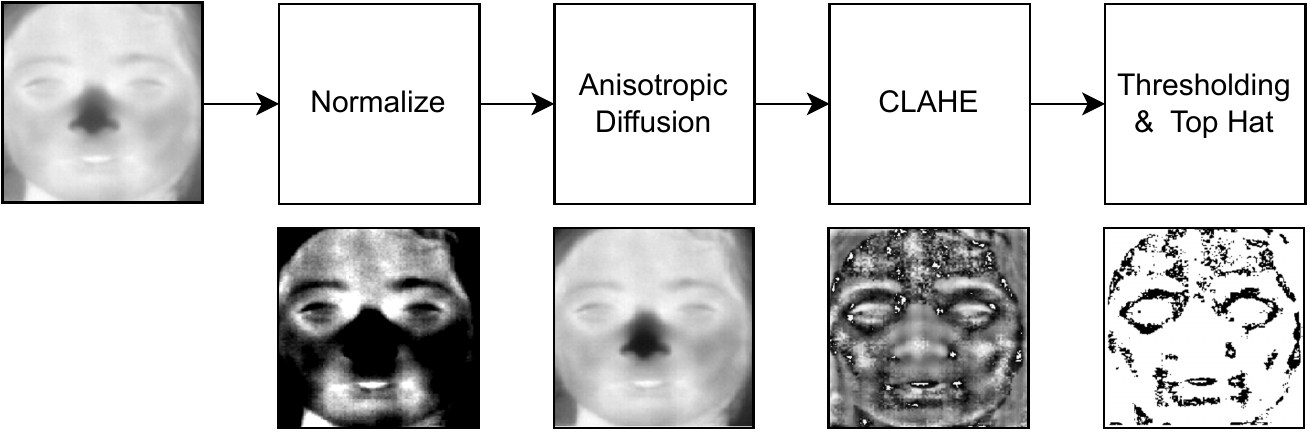}
        \caption{Vessel Extraction Processing}
        \label{aniso}
     \end{subfigure}
\caption{\textbf{Facial anatomy and Preprocessing for Vessel Maps.}}
\label{aso_proc}
\end{figure}

\subsection{Baselines}
First, we explore manual approaches using existing facial landmark algorithms in order to capture right and left eye coordinates implicit for affine matrix estimation. One algorithm is the Google MediaPipe \cite{lugaresi2019mediapipe} Face Mesh Active Appearance Model (AAM) based on 3D Morphable Models \cite{kartynnik2019real} that implements a Multi-task Cascaded Convolutional Network (MTCNN) model \cite{zhang2016joint} using an InceptionResnetV1 model pre-trained on VGGFace2. Each detected face returns an array of 468 points for three coordinates used as keypoints. For the Devcom dataset, the AAM fails to detect landmarks for 40\% of the thermal faces, thereby leaving only 60\% of facial pairs usable. For the remaining pairs, the scale of the desired registered image must be determined by estimating ratios between eye distances among the current and target image. This enables retrieval of geometric parameters to compute the affine matrix. No singular set of parameters can address all variations of warp, and as a result, they must be calculated manually for every VT pair. Samples results shown in Figure \ref{manual} demonstrate the imperfection of this manual approach. Since the manual pipeline is not feasible for multiple VT datasets that contain a variable level of warp, scale, and abrupt changes, we compare our approach against the Nemar model \cite{arar2020unsupervised} as the closest to Vista Morph's unsupervised approach. 


\subsection{Experimental Protocol} \emph{First}, we train Vista Morph and Nemar in two directions to: (1) align the thermal face relative to the visible face (T$\sim$V), and (2) for the Eurecom dataset, align visible relative to thermal (V$\sim$T). As a result, we perform a total of four VT facial registration experiments: (1) Devcom (T$\sim$V), (2) Carl (T$\sim$V), (3) Eurecom (V$\sim$T), (4) Eurecom (T$\sim$V). For Carl and Eurecom, we trained the T$\sim$V alignment, using $L_{FFT}$ loss due to No- and Low-Light images. \emph{Second}, we register the entire dataset (train, test) using Vista Morph and Nemar. \emph{Third}, we conduct a downstream generative task, for image-to-image translation using the VTF-GAN \cite{ordun2023visible}, a conditional GAN specifically designed for Visible-to-Thermal (V2T) facial image-to-image translation. We generate both visible and thermal faces using both the unregistered original data and the Vista Morph registered pairs. 

\subsection{Evaluation} To score registration results, we use Structural Similarity Index Measure (SSIM) and Normalized Cross Correlation (NCC) of the edge maps (e.g. morphological gradients of the visible and thermal images), in addition to Mutual Information (MI) \cite{kern2007robust} between both spectra. For generative results, we score with Frechet Inception distance (FID) \cite{heusel2017gans} and LPIPS \cite{zhang2018unreasonable}. Lastly, we analyze a sample of Devcom generated thermal faces for retention of identity through facial vasculature maps. 

\subsection{Implementation} For registration experiments, we train our model and the baseline using PyTorch, to 50 epochs with a batch size of 32. For generative experiments, we train the VTF-GAN and VTF-Diff from scratch, to 200 epochs with a batch size of 32. For all experiments, we use automatic mixed precision and parallel training on two RTX-8000 GPUs. Training Vista Morph is fast, where registration is learned approximately 1 hr. on the Devcom dataset.

\section{Results}

\subsection{Registration}
In Table \ref{Registration_Results}, Vista Morph outperforms the Nemar baseline for T$\sim$V alignment on all datasets. For Mutual Information, Vista Morph is comparable to Nemar with only a marginal difference (0.260, 0.279). Vista Morph shows the greatest registration gains  with the Carl dataset for all three metrics (5.1\%, 105.0\%, 26.8\%), which exhibits several Low-Light settings. Similarly, the T$\sim$V alignment for Eurecom improves using Vista Morph, as this dataset includes Low- and No-Light visual images. We show in Figure \ref{c} that Nemar fails to register T$\sim$V, since the visible face is captured in a No-Light setting. The T$\sim$V results in Figure \ref{a} and \ref{b}, show the precise alignment of Vista Morph despite hair texture and differences in scale and height. 

An intuitive view are difference maps shown in Figure \ref{difference}. These plots visualize the shift of pixels between the registered and original images. For example, the top row of Figure \ref{difference} shows the difference without registration where the thermal glasses (blue) are not aligned to the visible eyes (red). After registration with Vista Morph, the glasses are superimposed on the eyes, whereas the baseline still demonstrates misalignment. Most noticeable is the Eurecom T$\sim$V on the second to last row which shows no difference, only a light orange Nemar plot. This is because no thermal image was registered. Carl plots show blue ringing effects and shadows around the Nemar difference map indicating the imperfect alignment to the visible face's scale. 

\subsection{Generation}
Table \ref{GAN_Results} shows results for the generative Visible-to-Thermal (V2T) image-to-image translation tasks using registered and unregistered training data. In all cases, scores improve significantly after VT pairs are registered. For FID scores, Nemar achieves slightly lower scores than Vista Morph yet LPIPS scores are 6.8\% lower for Vista Morph. For the Carl and Eurecom V2T translations, their thermal faces show a -13.9\% and -12.4\% decrease in FID, and, -13.1\% and -12.4\% decrease in LPIPS, respectively. Sample generated images for the V2T direction are provided in Figure \ref{gan}. Upon qualitative inspection, when using unregistered data or Nemar-registered pairs, the generated thermal faces (``GAN:V2T", ``GAN:V2(Ne)T") introduce more artifacts, less texture and consistency, and similarity to their ground truth thermal faces. The ``GAN:V2(VM)T" faces retain distribution of pixel color (e.g. thereby thermal temperature), perceptual clarity, and hair texture which is important for maintaining minority and female identities. Additional results are in the Supplementary Materials.

\begin{table}[]
\centering
\begin{adjustbox}{width=0.48\textwidth}
\begin{tabular}{@{}lllllr@{}}
\toprule
              &     & \multicolumn{4}{c}{\textbf{FID $\downarrow$}}                          \\ \midrule
\textbf{Dataset} & \textbf{Translation} & \textbf{No Reg.} & \textbf{VM} & \textbf{Ne} & \multicolumn{1}{l}{\textbf{Diff.}} \\ \midrule
Devcom        & V2T & 60.357  & 50.740          & \textbf{50.338} & 0.8\%            \\
Carl          & V2T & 52.865  & \textbf{44.765} & 51.972          & \textbf{-13.9\%} \\
Eurecom (Thr) & V2T & 70.221  & \textbf{69.893} & 79.810          & \textbf{-12.4\%} \\ \midrule
\textbf{Dataset} & \textbf{Translation} & \multicolumn{4}{c}{\textbf{LPIPS $\downarrow$}}  \\ \midrule
Devcom        & V2T & 0.279   & \textbf{0.218}  & 0.234           & \textbf{-6.8\%}  \\
Carl          & V2T & 0.190   & \textbf{0.168}  & 0.193           & \textbf{-13.1\%} \\
Eurecom (Thr) & V2T & 0.157   & \textbf{0.144}  & 0.165           & \textbf{-12.4\%} \\ \bottomrule
\end{tabular}
\end{adjustbox}
\caption{\textbf{Generation Results for V2T Image Translation.} Best FID and LPIPS scores in \textbf{bold}. VM (Vista Morph), Ne (Nemar), Diff: Relative perc. change of VM over Ne.}.
\label{GAN_Results}
\end{table}

\begin{figure}[t]
    \centering
    \includegraphics[width=0.48\textwidth]{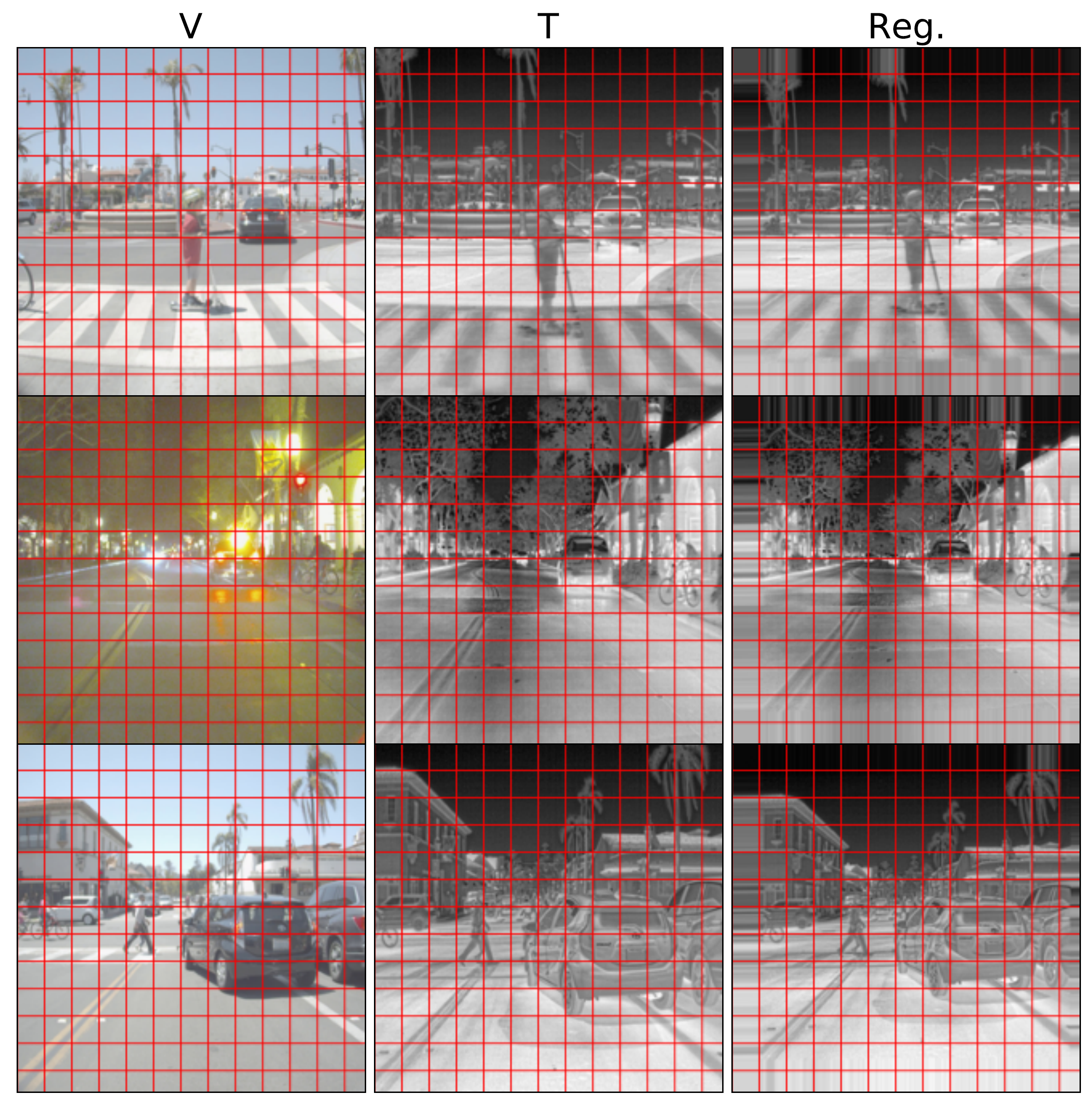}
    \caption{\textbf{Vista Morph registers Non-Facial Data - ADAS Street Scenes}. For scenes of pedestrians in hard light and low-light, with different scales of objects, Vista Morph aligns  $T \sim V$), shown in the last column.}
    \label{adas}
\end{figure}

\subsubsection{Generated Identity Analysis}
To visualize how Vista Morph registered data improves the generation of subject identity, we turn to facial vascular network extraction as defined by \cite{buddharaju2007physiology}. Building on biometric work stemming from retrieval of venous structure in palms and wrists \cite{francis2017novel,zhang2007palm}, Buddharaju, et al. \cite{buddharaju2007physiology} propose thermal vasculature as a unique biometric that can be extracted from thermal faces through basic image processing shown in Figure \ref{aniso}: anisotropic diffusion \cite{perona1990scale} to remove noise and enhance sigmoid edges, followed by CLAHE (Contrast Limited Adaptive Histogram Equalization) \cite{pizer1987adaptive}, and then finally top hat segmentation. We use samples from the Devcom dataset as a test bed since faces are close-up with minimal apparel. To measure similarity between the generated and real identity, we calculate the Peak Signal to Noise Ratio (PSNR) \cite{hore2010image} between vessel diagrams. Subjects in Figure \ref{vessels} show the facial vein, labial arteries (mouth and nose), angular artery and vein (eyes), superficial temporal artery and vein (edge of face), and supraorbital artery and vein (forehead). PSNR between the GAN is trained on registered pairs. For example, Subject 1 shows a PSNR of 11.018 of vessel maps before registration. Notice how ``G``, the generated image trained on unregistered data shows an identity dissimilar to its respective ground truth, ``T". The PSNR of their respective vessel maps is 11.018. When Vista Morph registers T$\sim$V in ``RT", the PSNR between vessel maps increases to 11.317, indicating that the generated subject's identity is more similar to the ground truth, when registered. Similar evidence can be seen in Subjects 2 and 3 that display extreme variance in scale as well as head tilt between VT pairs. 

\section{Ablation Studies}

\subsection{Architecture and Patch Size}
We conducted a brief ablation study using the Devcom dataset (T$\sim$V). When using the traditional U-NET for the STN, all scores decrease significantly when compared to our implementation (SSIM: -2.02\%, NCC: -104.15\%, MI: -19.03\%). Patch=32 size led to the worst results where the affine matrix could not be estimated. For facial images, more patches (e.g. increase of patch size) which preserve positional information, empirically leads to better registration results. 

\begin{table}[]
\centering
\begin{adjustbox}{width=0.45\textwidth}
\begin{tabular}{@{}llll@{}}
\toprule
\textbf{Ablation Exp.} & \textbf{SSIM Edges} & \textbf{NCC Edges} & \textbf{MI}    \\ \midrule
U-NET, Patch=64                  & 0.880               & -0.006             & 0.218          \\
Patch = 16             & 0.894               & 0.068              & 0.230          \\
Patch = 32             & N/A                 & -0.054             & 0.075          \\
Baseline             & \textbf{0.898}      & \textbf{0.147}     & \textbf{0.269} \\
Patch = 128            & 0.897               & 0.140              & 0.262         \\ \bottomrule
\end{tabular}
\end{adjustbox}
\caption{\textbf{Ablation Study for Devcom T$\sim$V}. Results are compared to the Vista Morph Baseline (ViT, Patch=64).}
\end{table}

\subsection{Non-Facial Domains}
Vista Morph successfully registers non facial pairs, namely the FLIR ADAS street scenes dataset. We train a ``deeper" STN regressor by adding two more linear layers to the baseline. We believe this enables finer focus of object features such as persons, stop signs, pedestrian crossings, and vehicles. Further, we incorporate Fourier Loss since ADAS includes several No- and Low-Light settings. Further, we find that 32 patches and removing the morphological loss also improves image quality. Registration scores for Vista Morph are 0.672 SSIM, 0.318 NCC, 0.607 MI compared to no registration at 0.627 SSIM, 0.250 NCC, 0.516 MI.

\subsection{Robustness}
\begin{figure}[t]
     \centering
    \begin{subfigure}[b]{0.48\textwidth}
        \centering
        \includegraphics[width=\textwidth]{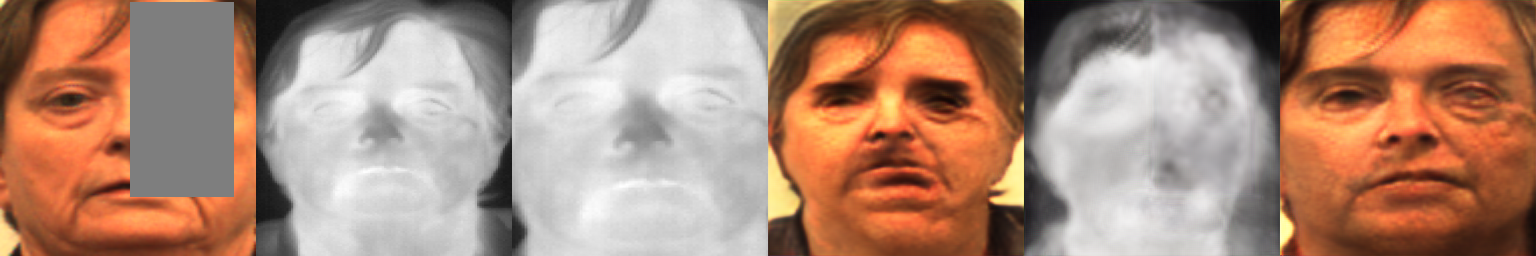}
        \caption{Erasure Visible}
        \label{ablation2-a}
     \end{subfigure}    
     \begin{subfigure}[b]{0.48\textwidth}
        \centering
         \includegraphics[width=\textwidth]{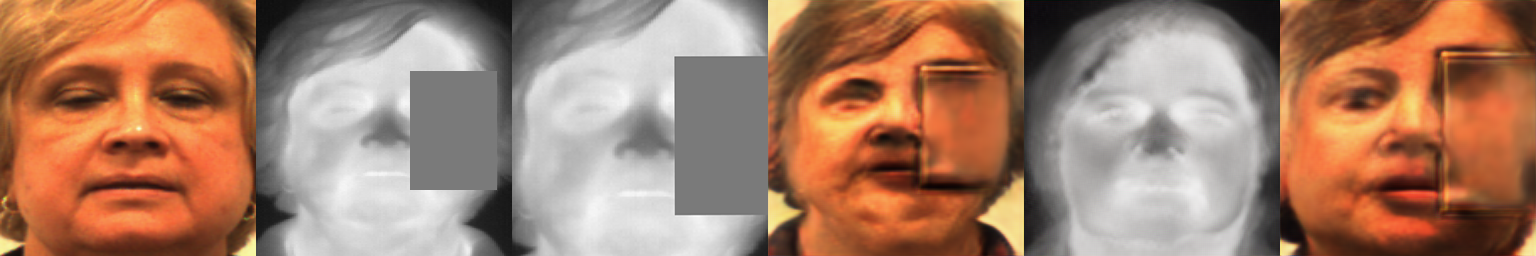}
         \caption{Erasure Thermal}
         \label{ablation2-b}
     \end{subfigure}
     \begin{subfigure}[b]{0.48\textwidth}
        \centering
        \includegraphics[width=\textwidth]{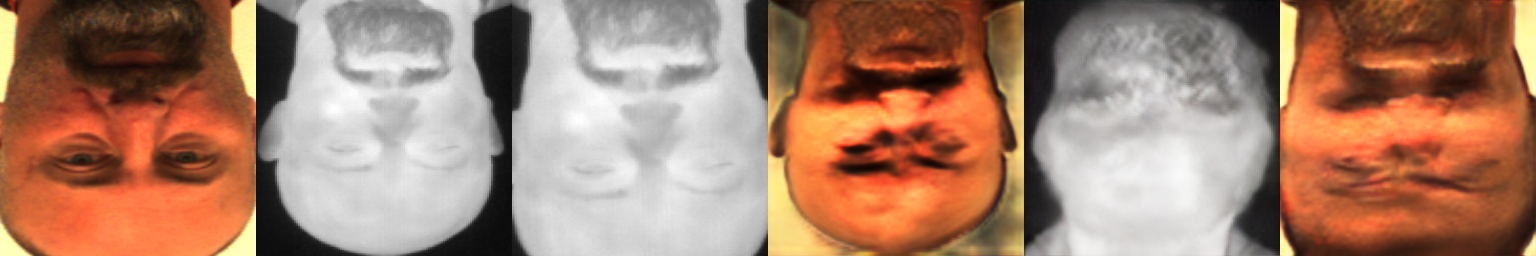}
        \caption{Vertical Flip}
        \label{ablation2-c}
     \end{subfigure}
    \caption{\textbf{Vista Morph robustness against random geometric transformations and erasure}. Columns left to right: Real Visible, Real Thermal, Registered Thermal, A1, Fake Thermal, A2}
    \label{abl2}
\end{figure}

To illustrate Vista Morph's robustness, we test registration with geometric transformations and random erasure at inference time on the input pairs. Despite the permutations, Vista Morph successfully registers the T$\sim$V. In Figure \ref{ablation2-a}, the thermal image registers to the scale of the visible input regardless of erasure. Figure \ref{ablation2-b} shows the expected behavior, that the entire thermal image is registered. Here, the generated $\hat{A_1}$ and $\hat{A_2}$ images demonstrate the translation of the erasure pixels. Figure \ref{ablation2-c} shows that when vertically flipped, Vista Morph will register the thermal image accordingly with respect to the visible geometry.

\begin{table}[]
\centering
\begin{adjustbox}{width=0.48\textwidth}
\begin{tabular}{@{}llllll@{}}
\toprule
            & \textbf{SSIM} $\uparrow$ & \textbf{NCC} $\uparrow$ & \textbf{MI} $\uparrow$   & \textbf{VTF-GAN} $\downarrow$ & \textbf{VTF-Diff} $\downarrow$ \\ \midrule
No Reg.     & 0.833      & 0.285     & 0.493 & 111.142 & \textbf{67.078}   \\
Nemar       & \textbf{0.843}      & 0.217     & \textbf{0.556} & \textbf{83.346}  & 81.398   \\
Vista Morph & 0.832      & \textbf{0.281}     & 0.530 & 87.887  & 86.102   \\ \bottomrule
\end{tabular}
\end{adjustbox}
\caption{\textbf{Limitations of $V \sim T$ Registration and T2V Generation.} Vista Morph underperforms for visible face registration and the T2V translation when using the Vista Morph registered data. FID scores are provided for the VTF-GAN and -Diff model results. SSIM, NCC, MI are the same as Table \ref{Registration_Results}, abbreviated for space.}
\label{Vis-Limits}
\end{table}

\section{Limitations}
Unfortunately, for $V \sim T$ registration, Vista Morph underperforms. To explore how registration effects a different generative model, we trained a conditional Denoising Diffusion Probabilistic Model (DDPM) \cite{ho2020denoising,nichol2021improved} called VTF-Diff \cite{ordun2023visible} on registered and non-registered images in the T2V direction. GAN results improve with registration (Nemar), but registration does not improve the Diffusion results. Although the generated diffusion results look geometrically and perceptually similar, there are differences in skin color, eye detail, and clothing. Further tests are needed beyond the Eurecom dataset. 

\begin{figure}[ht]
\centering
     \begin{subfigure}[b]{0.48\textwidth}
        \centering
         \includegraphics[width=\textwidth]{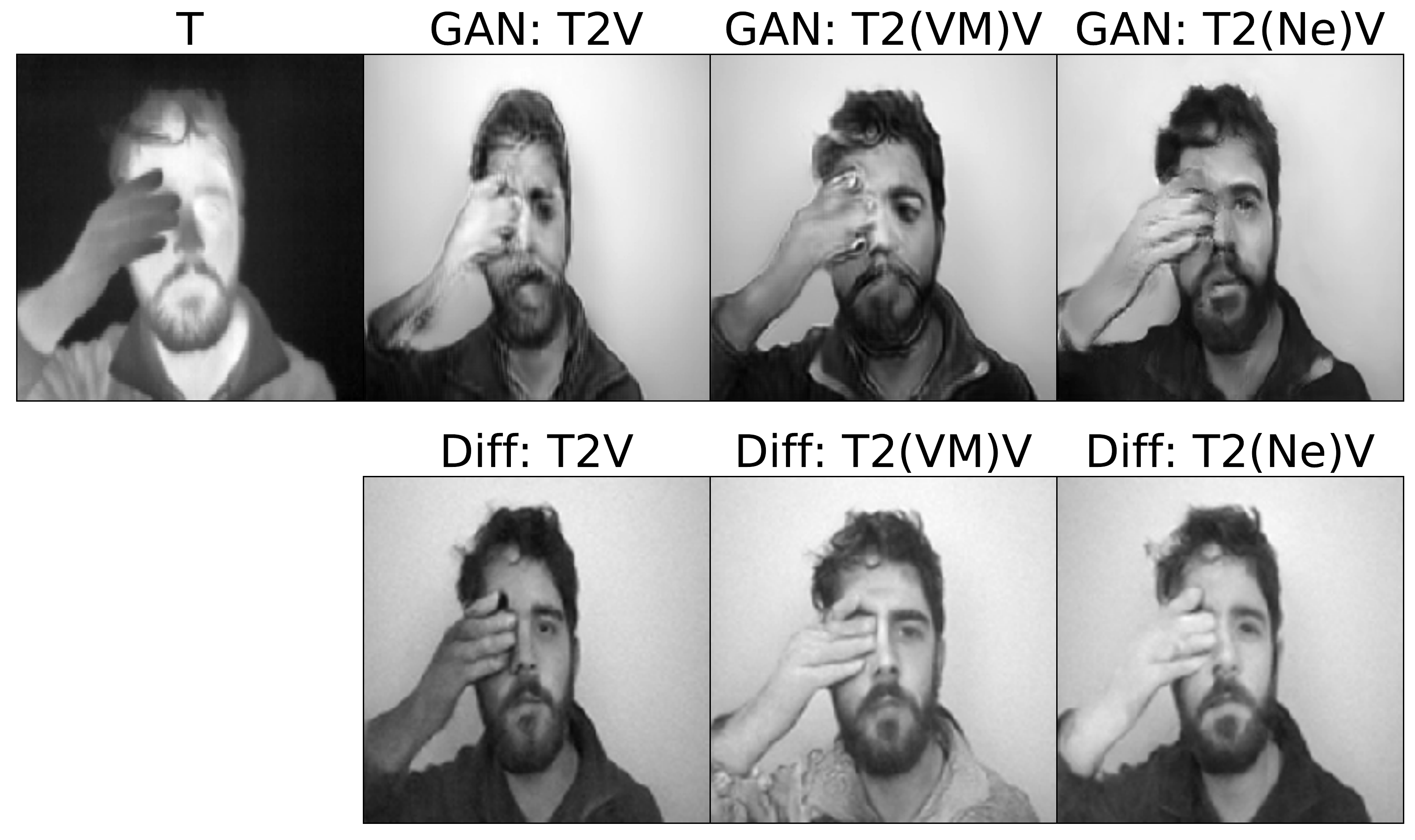}
     \end{subfigure}  
    \caption{\textbf{Generated T2V Image Translation Sample.} Training Vista Morph $V \sim T$ aligned images does not improve GAN results. Diffusion results quantitatively perform better without registration, despite looking qualitatively similar.}
    \label{ddpm}
\end{figure}

\section{Conclusion}
We present Vista Morph, the first unsupervised VT facial registration model that aligns facial pairs without a reference or feature matching. We evaluate three VT facial datasets leading to significantly improved registration results over state-of-the-art methods. Further, we show that image quality from a generative Visible-to-Thermal image translation task improves with regards to perceptual clarity and identity when training a GAN on Vista Morph registered pairs. We support our findings with thermal vessel maps and demonstrate Vista Morph can register non-facial domains. Future work includes assessment of generated thermal faces by thermal specialists and user studies, and finer investigation into the consistency of results for diverse demographic samples.

\def\bibfont{\fontsize{9}{10}\selectfont}
\bibliographystyle{IEEEbib}
\bibliography{references}

\end{document}